\pdfoutput=1

\documentclass[11pt]{article}
\PassOptionsToPackage{hyphens}{url}

\usepackage[]{acl}

\usepackage{caption}
\usepackage{subcaption}
\usepackage{svg}
\usepackage{times}
\usepackage{latexsym}
\usepackage[shortlabels]{enumitem}

\usepackage{amsmath}
\usepackage{algorithm}
\usepackage{algorithmicx}
\usepackage{algpseudocode}

\DeclareMathOperator{\Mode}{mode}

\usepackage[T1]{fontenc}

\usepackage[utf8]{inputenc}

\usepackage{microtype}

\usepackage{inconsolata}

\usepackage{makecell}

\usepackage{booktabs}
\usepackage{todonotes}
\usepackage{tabularray}
\usepackage{ninecolors}
\graphicspath{{imgs}}
\usepackage{multirow}
\usepackage{enumitem}
\usepackage{xspace}

\UseTblrLibrary{booktabs}
\usepackage{newfloat}
\usepackage{listings}

\definecolor{cBLUE}{HTML}{3282B8}
\definecolor{cGREEN}{HTML}{60A561}
\definecolor{cORANGE}{HTML}{FA824C}
\definecolor{cYELLOW}{HTML}{f0C808}
\definecolor{cLightGrey}{HTML}{CECECE}

\definecolor{cRED}{HTML}{ED1B23}

\definecolor{cNatureORANGE}{HTML}{EAAA60}
\definecolor{cNatureRED}{HTML}{E68B81}
\definecolor{cNaturePURPLE}{HTML}{B7B2D0}
\definecolor{cNatureBLUE}{HTML}{7DA6C6}
\definecolor{cNatureGREEN}{HTML}{84C3B7}

\usepackage{graphicx} 
\title{More Samples or More Prompts? Exploring Effective Few-Shot In-Context Learning for LLMs with In-Context Sampling}

\author{Bingsheng Yao  \\ Rensselaer Polytechnic Institute \\ Northeastern University \\
\And Guiming Chen \\ The Chinese University \\ of Hong Kong, Shenzhen \\
\And Ruishi Zou \\ Tongji University \\
\AND Yuxuan Lu \\ Northeastern University \\
\And Jiachen Li \\ Northeastern University \\
\And Shao Zhang  \\ Shanghai Jiao Tong University \\ 
\AND Yisi Sang \\ Northeastern University \\
\And Sijia Liu \\ Michigan State University \\
\And James Hendler \\ Rensselaer Polytechnic Institute \\
\AND Dakuo Wang  \thanks{~~Corresponding Author: \href{mailto:d.wang@northeastern.edu}{d.wang@northeastern.edu}. This work was done while Guiming, Ruishi, and Shao were visiting students at Northeastern University.} \\ Northeastern University \\
}

\begin{document}
\maketitle
\begin{abstract}


While most existing works on LLM prompting techniques focus only on how to select a better set of data samples inside one single prompt input (In-Context Learning or \textbf{ICL}), why can not we design and leverage multiple prompts together to further improve the LLM's performance?
In this work, we propose In-Context Sampling (\textbf{ICS}), a low-resource LLM prompting technique to produce confident predictions by optimizing the construction of multiple ICL prompt inputs. 
Extensive experiments with three open-source LLMs (FlanT5-XL,  Mistral-7B, and Mixtral-8x7B) on four NLI datasets (e-SNLI, Multi-NLI, ANLI, and Contract-NLI) and one QA dataset (CommonsenseQA) illustrate that ICS can consistently enhance LLMs'  performance.
An in-depth evaluation with three data similarity-based ICS strategies suggests that these strategies can further elevate LLM's performance, which sheds light on a new yet promising future research direction. 

\end{abstract}

\section{Introduction}
\label{sec:intro}

Large Language Models (LLMs) with billions of parameters, such as FLAN-T5~\cite{chung2022scaling}, LLaMA~\cite{touvron2023llama, touvron2023llama2}, and Mistral~\cite{jiang2023mistral}, have demonstrated exceptional natural language interpretation capability in terms of understanding versatile prompt inputs\footnote{We use ``prompt input'' to refer to the composition of prompt structures, including the task narrative instructions, plus in-context examples, and the targeting data for inference.}. 
In comparison with much smaller language models like BERT~\cite{devlin2018bert} and GPT~\cite{radford2018improving}, such LLMs can understand not only more complex and detailed task narratives but also a few task examples with annotations within the prompt inputs, namely few-shot In-Context Learning (ICL)~\cite{brownLanguageModelsAre2020, shin2022effect}.

\begin{figure}[!t]
\centering
\includegraphics[width=.98\columnwidth]{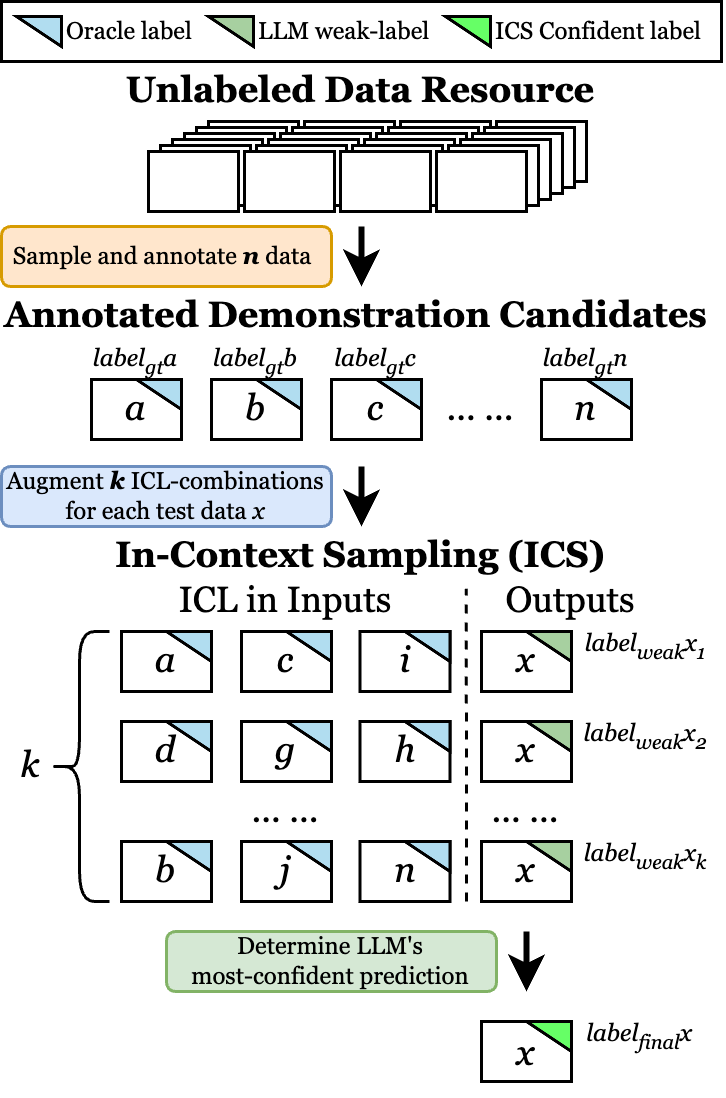}
\caption{Our proposed ICS paradigm comprises three steps: 1) \textbf{sample} representative ICL demonstration candidates, 2) \textbf{augment} different ICL prompt inputs from the sampled candidates and \textbf{acquire} LLM's prediction for each input correspondingly, and 3) \textbf{determine} and \textbf{vote} LLM's most confident prediction.}
\vspace{-1.5em} 
\label{fig:ics_overview}

\end{figure}

As a prominent prompting strategy to exploit LLMs' task-solving capabilities, especially for unseen tasks, ICL inserts a few data examples as well as their corresponding annotations into the prompt input. 
The data examples, along with their annotations, serve as demonstrations\footnote{ We use ``examples'' and ``demonstrations'' interchangeably to refer to the few-shot data examples within the prompts.} for the targeting task. 
The demonstrations are expected to facilitate LLMs' better understanding of the task narrative, the expected outputs, and potentially the underlying rationales needed for solving the task. 
Several recent works investigate the influence of different ICL setups, including the number, ordering, and combinations of demonstrations~\cite{wang2022iteratively, luFantasticallyOrderedPrompts2022, yoo2022ground}. 
However, there is no common ground for the best ICL strategy yet.

Moreover, despite LLMs' superb natural language interpretation and generation capability, real-world tasks requiring extensive domain expertise remain challenging for LLMs (e.g., children's education and mental issue detection~\cite{chen2023fairytalecqa, xu2023mentalllm,zhang2023rethinking}), and thus, how to exploit LLMs' ability with ICL for solving these tasks is an under-explored topic but holds great promise.

We hypothesize that different ICL demonstrations provide LLMs with distinct knowledge about the task, leading to disparate understanding and predictions for the same data.  
Consequently, a research question emerges: 
\textbf{Can we augment multiple ICL prompt inputs efficiently to facilitate more accurate and confident LLM predictions?}

To address this question, we propose \textbf{In-Context Sampling (ICS)}, a low-resource methodology inspired by the \textit{query-by-committee} strategy~\cite{seung1992query, liere1997active} and the \textit{few-shot In-Context Learning} approach. ICS follows a three-step pipeline as shown in Figure~\ref{fig:ics_overview}: 
\begin{enumerate}[nosep]
\item \textbf{Sample} demonstration candidates;
\item \textbf{Augment} ICL prompt inputs and predictions;
\item \textbf{Vote} the most confident label.
\end{enumerate}
We also propose three data similarity-based ICS strategies inspired by established data sampling strategies for Active Learning~\cite{settles2009active}.
We believe ICS can be a more reliable prompting paradigm than the traditional ICL, better squeezing LLM's task-solving capabilities and seamlessly supporting ``plug-and-play'' customizations.

Our evaluation of the ICS paradigm comprises bi-fold.
First, we benchmark the effectiveness of a baseline ICS strategy with the traditional ICL approach on \textbf{three} open-source LLMs (FLAN-T5-XL~\cite{chung2022scaling}, Mistral-7B~\cite{jiang2023mistral}, and Mixtral-8x7B~\cite{jiang2024mixtral})\footnote{We also experiment with Llama2~\cite{touvron2023llama2} and discussed its limited performance in Appendix~\ref{app:llama2}.} over \textbf{five} datasets, including four natural language inference (NLI)~\cite{bowman2015large} datasets as well as the CommonsenseQA (CQA) dataset~\cite{talmor2018commonsenseqa}.
Among four NLI datasets, three are general-domain NLI tasks of increasing difficulty (e-SNLI~\cite{camburu2018snli}, Multi-NLI~\cite{williams2017broad}, and ANLI~\cite{nie2019adversarial}), and the last one is Contract-NLI~\cite{koreeda-manning-2021-contractnli-dataset}, a domain-specific NLI dataset for the real-world contract review task.
We also investigate how different sample sizes and the number of ICL prompt inputs affect model reliability in terms of performance enhancement.
Results indicate that ICS can consistently improve prediction accuracy and robustness despite LLMs demonstrating different levels of ICL capabilities.

We further investigate the additional advantages provided by three different ICS strategies through simulations with the best-performing setting from the previous experiment, compared with the random ICS and traditional ICL approaches on the aforementioned four datasets.
Despite being conceptually straightforward, all three types of data-based strategies can effectively and consistently improve LLM performance, leading to a broader research scope to exploit ICS in the future.

\section{Related Work}
\label{sec:related}



\subsection{Large Language Models}
 
Large Language Models (LLMs)~\cite{brownLanguageModelsAre2020,touvronLLaMAOpenEfficient2023,touvron2023llama2,openai2023gpt4} show impressive capability in understanding free-form instructions and generating high-quality content in a variety of tasks~\cite{weiFinetunedLanguageModels2021, sanh2021multitask, chung2022scaling, mahmood2023llmpowered, yao2023human, yang2024talk2care}. 
For instance, \citet{weiFinetunedLanguageModels2021} proposed FLAN-T5, a model trained to follow natural language instruction on over 60 NLP tasks.
\citet{ouyang2022training} proposed a pipeline to instruction-finetune LLM with Reinforcement Learning from Human Feedback.
In addition, various prompting methods such as Chain-of-Thoughts~\cite{weiChainofThoughtPromptingElicits2023, chung2022scaling} and In-Context Learning (ICL)~\cite{brownLanguageModelsAre2020} have been developed to exploit LLMs' potential, where the former technique asks models to generate a sequence of rationales, and the latter methodology allows LLMs to learn from few-shot examples in the input context.
Our ICS paradigm extends the traditional ICL approach to improve the performance and confidentiality of LLM predictions.

\subsection{In-Context Learning Optimization}
Optimizing ICL performance has garnered significant attention recently.
\citet{dongSurveyIncontextLearning2023} summarized three categories for different ICL optimization approaches: fine-tuning with ICL, ICL sample selection, and analyzing order sensitivity.
Finetuning with ICL generally requires a significant amount of computing resources and effort to tune model parameters, such that \citet{weiFinetunedLanguageModels2021} proposed an instruction tuning method that improves both zero-shot and few-shot In-Context Learning performance.
Sample selection in ICL has been demonstrated to have a considerable impact on model performance \cite{zhangActiveExampleSelection2022,rubinLearningRetrievePrompts2022,liUnifiedDemonstrationRetriever2023}.
\citet{zhangActiveExampleSelection2022} initiated a reinforcement learning technique to select more advantageous samples for in-context demonstration.
\citet{rubinLearningRetrievePrompts2022} proposed a two-staged method with an unsupervised retriever followed by a supervised model. 
Some work focused on reducing LLM's ICL order sensitivity issue. \citet{luFantasticallyOrderedPrompts2022} proposed multiple sample sorting methods, while \citet{liuWhatMakesGood2022} introduced a method for arranging examples based on their semantic similarity.
A few other works attempted to exploit the benefits of the ICL pipeline to improve model performance, better alignment, and minimize reliance on external demonstrations~\cite{yu-2023-cold, linUnlockingSpellBase2023, kimSelfGeneratedInContextLearning2022}.

\subsection{Sampling Strategies}
The data sampling strategy is a key element of many low-resource learning paradigms that attempt to select the most representative examples, such as Active Learning (AL)~\cite{settles2009active}. 
Following established works, the data sampling strategies have been mainly categorized into three categories: \emph{model-based}, \emph{data-based}, and \emph{hybrid}~\cite{settles2009active, olsson2009literature, fu2013survey, schroder2020survey, ren2021survey, zhang-etal-2022-survey, schroder-etal-2022-revisiting, lu2023human}.

Model-based strategies aim to find the data with the most model uncertainty~\cite{wang2017active, zeng-etal-2019-empirical}. 
For instance, \citet{margatina-etal-2021-active} and \citet{zhang-etal-2022-allsh} explored using the divergence of a model's prediction as a measurement of model uncertainty. 
Data-based strategies, on the other hand, aim to find the most diverse or representative data in the data space~\cite{erdmann-etal-2019-practical, prabhu-etal-2019-sampling, karamcheti-etal-2021-mind}. 
Such that \citet{deng2018adversarial, sinha2019variational} leveraged adversarial learning to select the most representative data.
In contrast to model-based strategies, data-based strategies are generally model-agnostic and demand fewer computational resources but necessitate the analysis of unlabeled samples.
Hybrid or ensemble Sampling Strategies integrate various strategy types in unison~\cite{krogh1994neural, tang-etal-2002-active, melville2004diverse, donmez2007dual, zhu-etal-2008-active, ambati-etal-2011-multi}. 
For instance, \citet{qian-etal-2020-learning} proposed a combined approach of a diversity-based and an uncertainty-based tactic to benefit from both strategies.

\section{ICS Prompting Paradigm}
\label{sec:ics}

Given a natural language task instruction $I$ and a datum to predict $x\in \mathcal{D}$, 
LLMs can take the In-Context Learning (ICL) input format, denoted as:
\begin{equation}
    \{I+(x^{icl}_{1},y^{icl}_{1})+...+(x^{icl}_{m},y^{icl}_{m})+x\}
\end{equation}
where $(x^{icl}_{m},y^{icl}_{m})$ denotes an oracle-annotated in-context demonstration. 
We believe in-context demonstrations can provide LLMs with two types of knowledge: 1) \textbf{explicit} insights to interpret the task instruction $I$ and expected outputs through $(y^{icl}_{1},...,y^{icl}_{m})$ and 2) \textbf{implicit} guidance for how to solve the task via demonstrations $(x^{icl}_{m} \rightarrow y^{icl}_{m})$.
We hypothesize that \textbf{different sets of ICL demonstrations provide LLMs with disparate implicit knowledge about the task}; thus, LLMs may alter their predictions for the same data $x$ given different ICL prompt inputs, but the predictions will eventually converge to a most confident result.

Our hypothesis stands on the shoulder of the \textit{query-by-committee}~\cite{seung1992query, liere1997active} strategy that has been around for a long time. 
The original concept is to ask a committee of models to vote on whether the unlabeled data needs to be annotated, where the voting models focus on competing hypotheses.
However, most existing works focused on measuring the disagreements among committee models~\cite{engelson-dagan-1996-minimizing, mccallum1998employing} and creating different committees with probabilistic and non-probabilistic models~\cite{dagan1995committee, freund1997decision, 10.5555/645527.657478, melville2004diverse, tomanek2009reducing, 10.1145/775047.775087}.

In this work, we present \textbf{In-Context Sampling (ICS)}, a low-resource paradigm for LLMs through effectively augmenting ICL prompt inputs, as shown in Figure~\ref{fig:ics_overview}. 
We view the ICS strategy as exploring efficient approaches to create committee ICL prompt inputs and query LLMs for the most confident prediction.
ICS consists of three steps: 
\begin{enumerate}[nosep]
\item \textbf{Sample} demonstration candidates and acquiring oracle annotations,
\item \textbf{Augment} prompt inputs and label predictions with different ICL combinations, and
\item \textbf{Vote} the most confident label as the final prediction from augmented labels. 
\end{enumerate}

Before diving deep into the details of each step in ICS, we want to emphasize that our prototyped ICS strategies in this work are model-agnostic.
We will demonstrate the consistent effectiveness of a random baseline ICS strategy over the traditional ICL approach across five datasets and three LLMs in Section~\ref{sec:evaluation-benchmark}. 
More importantly, our ICS supports \textbf{``plug-and-play''} customizations by switching to different sampling, augmenting, and voting strategies with minimum effort. 
In addition to justifying the effectiveness of our proposed ICS pipeline and investigating the influence of different factors on performance improvement and robustness, we propose three types of model-agnostic ICS strategies and demonstrate their further improvements over the random ICS pipeline in Section~\ref{sec:evaluation-strategy}.
The following sections illustrate each ICS step in detail as well as our proposed three data similarity-based ICS strategies: diversity, similarity, and hybrid.
We also leave a broad research area to explore strategy variations in future work.

\subsection{Demonstration Candidate Sampling}
\label{sec:ics-sampling}

How to effectively select unlabeled examples to benefit model performance shares the same spirit as the Active Learning (AL) data sampling strategy~\cite{settles2009active}, where an AL strategy iteratively samples few examples for annotation and fine-tuning the model. 
The AL strategies are often categorized into three types, as illustrated above in Section~\ref{sec:related}: data diversity-based, model probability-based, and hybrid strategies.  
Existing work stated that the effectiveness of model-based strategies might differ from model to model~\cite{yao2023labels}, which could introduce irreverent factors when we benchmark our ICS versus the traditional ICL approach. 
In this work, we implement three different data similarity-based, model-agnostic strategies for ICS and evaluate their effectiveness in Section~\ref{sec:evaluation-strategy}, in addition to the baseline \textbf{Random} strategy where we demonstrate the effectiveness compared with traditional ICL approach in Section~\ref{sec:evaluation-benchmark}.
The mathematical notations of our proposed strategies are illustrated in Algorithm~\ref{alg:strategies}.

\paragraph{Diversity}
This strategy adheres to established cluster-based strategies (i.e., core-set)~\cite{sener2017active, yao2023labels}, aiming to identify examples \textbf{representative of all unlabeled data while maximizing the diversity among these selected instances}.
The concept of ensuring data diversity derives from the established \textit{density-weighted} sampling strategies~\cite{settles2008analysis, shen-etal-2004-multi}.
They assume the instances that can provide the most helpfulness should be the ones that are representative of the input space~\cite{he2023icld3ie}. 
In other words, the diversity among selected data should be maximized.
Specifically, our strategy calculates the cosine similarity for each data $x_i$, encoded with sentence-transformer~\cite{reimers-2019-sentence-bert}, with the following formula, where $\mathrm{embed}$ represents sentence-transformer embedding: 
\begin{equation}
\label{alg:avg-similarity}
\resizebox{\columnwidth}{!}{
    $s(x, \mathcal{D})=\cos\left(\mathrm{embed}(x), \frac{1}{|\mathcal{D}|}\sum_{j=1}^{|\mathcal{D}|} \mathrm{embed}(x_j)\right)$
}
\end{equation}
Subsequently, we rank the data by similarity score and retrieve $n$ examples with the same interval, ensuring the sampling diversity.
for instance, to sample 4 demonstrations from $10$ ranked unlabeled data, we choose the $1^{st}$, $4^{th}$, $7^{th}$, and $10^{th}$ data.

\begin{algorithm}[!t]

\caption{Proposed Data-based ICS Strategies}
\label{alg:strategies}

    \begin{algorithmic}[1]

    \Function{ics\_strategy}{$D, n, strategy$}  \Comment{$D:$ array of data content; $n:$ sample size; $strategy:$ strategy type }
        \State $A \gets (s(D_i, D))_{i \in \left[1, |D|\right]}$ \Comment{Average score}
        \State $S \gets argsort(A)$ \Comment{Descending order}

        \If {$strategy=$ \texttt{``diversity''}}
            \State $t=\left\lfloor \frac{\left|D\right|}{n} \right\rfloor$ \Comment{Step}
            \State \textbf{Return} $(S_{i})_{\substack{i \equiv 0 (\text{mod } t) \\ 1 \le i \le \left|D\right|}}$ 
        \ElsIf {$strategy=$ \texttt{``similarity''}}
            \State \textbf{Return} $(S_i)_{i \in [1, n) }$ 
        \ElsIf {$strategy=$ \texttt{``hybrid''}}
            \State $t=\left\lfloor \frac{\left|D\right|}{(n/2)} \right\rfloor$
            \State $R_{div} = (S_{i})_{\substack{i \equiv 0 (\text{mod } t) \\ 1 \le i \le \left|D\right|}}$ 
            \State $S' = S \ominus R_{div}$ \Comment{Array subtract.}
            \State $R_{sim} = (S'_i)_{i \in [1, n/2) }$
            \State \textbf{Return} $R_{div} \oplus R_{sim}$ \Comment{Array concat.}
        \EndIf
        
    \EndFunction
    
    \end{algorithmic}

\end{algorithm}

\paragraph{Similarity}
The similarity strategy shares the same procedure as the diversity strategy of calculating the averaged similarity score for each unlabeled data.
Nevertheless, the similarity strategy aims to find examples that are \textbf{of the highest averaged similarity to the whole unlabeled training data space} so that the sampled data will most likely be similar to the actual testing data. 
The underlying concept of this strategy is analogous to a family of \textit{density-weighted} sampling strategies that look for the ones that appear most in the unlabeled data space or are most similar to unlabeled data~\cite{fujii1999selective, xu2003representative, haffari-sarkar-2009-active}.
We follow the same mathematical procedure~\ref{alg:avg-similarity} above to calculate and rank the unlabeled data by the averaged similarity score. 
Then, differing from the diversity strategy, we retrieve $n$ highest-ranked examples from the ranked list.

\paragraph{Hybrid} Similar to the aforementioned line of ensemble strategies that incorporate different strategies altogether in Section~\ref{sec:related}, our hybrid strategy expects to benefit from both above-mentioned strategies, which aims to locate examples that are either representative of the sampling space or of the highest similarity to the whole space.
Subsequently, this hybrid strategy comprises two steps: first, sample $n/2$ examples following the diversity strategy, then sample $n/2$ examples following the similarity strategy from the remaining list.

\subsection{ICL Prompt Inputs Augmentation}
\label{sec:ics-augmentation}

As described in Section~\ref{sec:ics} and shown in Figure~\ref{fig:ics_overview} above, 
ICS augments label predictions for the same data by constructing multiple disparate ICL combinations from the demonstration candidates sampled in the previous step.
Many recent works~\cite{chen2023demonstrations,levy2023diverse,zhangActiveExampleSelection2022,rubinLearningRetrievePrompts2022,nguyen2023incontext,luFantasticallyOrderedPrompts2022,liuWhatMakesGood2022} attempted different ICL constructions by altering the demonstrations' numbers, orderings, prompts, or sampling strategies.
Nevertheless, there is no commonly recognized best strategy yet, and we believe models will learn disparate implicit guidance for solving the task via different demonstrations.
In this work, we utilize four NLI datasets of varying difficulties and fix \textbf{three} as the number of demonstrations per prompt input, consistent with the number of NLI categories. This setting also applies to the CQA task in our evaluation.

Still, the computation could be massive if we permutate every combination of the candidates. 
for example, $50$ demonstration candidates can result in $19,600$ 3-demonstration ICL combinations.
We believe, however, that ICS does not need every ICL combination to find the model's most confident label. 
Analogous to the \textit{query-by-committee} concept, where a few representative committee models vote for the best prediction, we plan to investigate a reasonable amount of ``committees'' (i.e., prompt inputs) that balance between establishing robust and reliable predictions and minimizing costs (i.e., computational resources, time, annotation efforts.

The task of augmenting ICL prompt inputs can be naturally viewed as a variation of the candidate sampling task for the previous step, where the underlying concept for both steps attempts to sample a few examples that could be potentially helpful to LLMs. 
Despite that, the optimal strategy for candidate sampling may not be optimal for augmenting prompt inputs in terms of effectiveness and helpfulness. 
The demonstrations in each prompt input are ordered in the same order as they are sampled.
In this work, we benchmark ICS over traditional ICL with a random strategy for augmenting prompt inputs in Section~\ref{sec:evaluation-benchmark}.
Analogous to the sampling step, we implement and evaluate three similarity-based, model-agnostic strategies proposed in Section~\ref{sec:evaluation-strategy} to select demonstrations for each prompt input. 
Specifically, for each data to be predicted, we iteratively sample three demonstrations from the candidate list with a certain strategy for $k$ times, remove them from the list, construct $k$ different prompt inputs, and thus, acquire $k$ predicted labels. 
For ICS strategy evaluation in Section~\ref{sec:evaluation-strategy}, we leverage the best-performing parameters from the benchmark experiment, where $n$=100 and $k$=10.

\subsection{Confident Prediction Voting}
\label{sec:ics-verification}

Once we acquire a set of predicted labels from the abovementioned ICS steps for each datum to be predicted, we can apply different voting algorithms to find LLM's most confident prediction. 
A straightforward design could be a majority vote algorithm to select the prediction with the most appearances among all the predictions for the current data, which is analogous to finding the mode value mathematically: $y^{final}=\Mode(y^{ics}_{1}, ..., y^{ics}_{k})$, where $y^{ics}_{k}$ denotes the prediction for each augmented prompt input of data $x$. 
In this work, we leverage the majority vote algorithm in our prototyped ICS pipelines.
We can further consider the model's different prediction confidences for a more complex algorithm design.
Additionally, we can envision ICS to \textbf{provide reliable unsupervised labels} to iteratively fine-tune LLM and compact models in resource-deficient scenarios where expert annotations are difficult and expensive to access.

\section{Evaluations}
\label{sec:evaluation}

The evaluation of our proposed ICS paradigm comprises bi-fold.
First, in Section~\ref{sec:evaluation-benchmark}, we execute a benchmark experiment between the random ICS strategy and traditional ICL approach on five datasets with two LLMs to demonstrate the paradigm effectiveness.
Additionally, we attempt to identify a sample size and the amount of augmented ICL combinations that strike a balance across three perspectives: 1) encompass sufficient diversity to represent the underlying data adequately, 2) possess robustness toward confident predictions, and 3) minimize annotation costs.
Subsequently, in Section~\ref{sec:evaluation-strategy}, we pick the best-performing parameters from the first experiment to compare the additional advantages of the three proposed ICS strategies described above in Section~\ref{sec:ics-sampling}.

\begin{figure*}[!t]
    \centering

    \begin{subfigure}{0.495\textwidth}
        \includegraphics[width=\columnwidth]{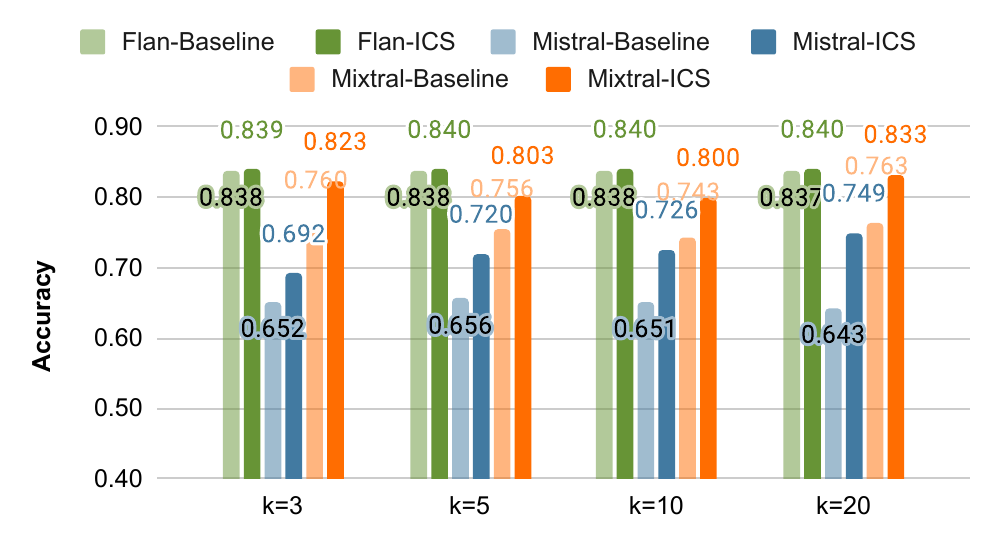}
        \caption{e-SNLI~\cite{camburu2018snli}}
    \end{subfigure}
    \hfill
    \begin{subfigure}{0.495\textwidth}
        \includegraphics[width=\columnwidth]{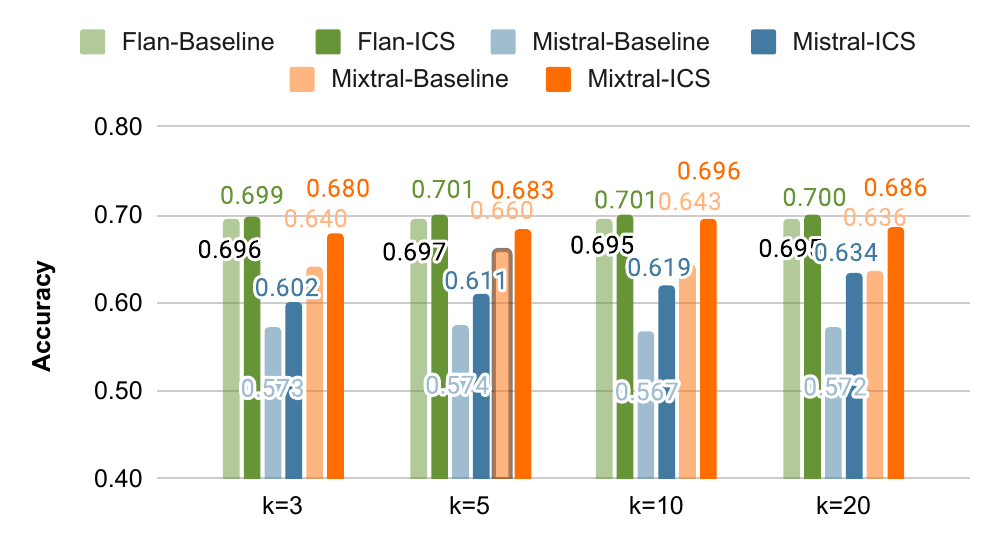}
        \caption{Multi-NLI~\cite{williams2017broad}}
    \end{subfigure}

    \begin{subfigure}{0.495\textwidth}
        \includegraphics[width=\columnwidth]{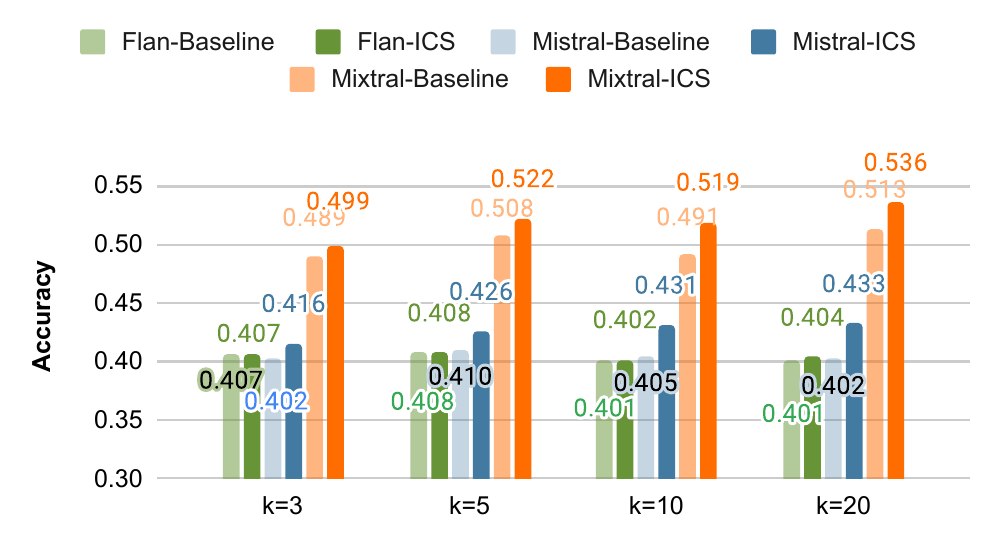}
        \caption{ANLI~\cite{nie2019adversarial}}
    \end{subfigure}
    \hfill
    \begin{subfigure}{0.495\textwidth}
        \includegraphics[width=\columnwidth]{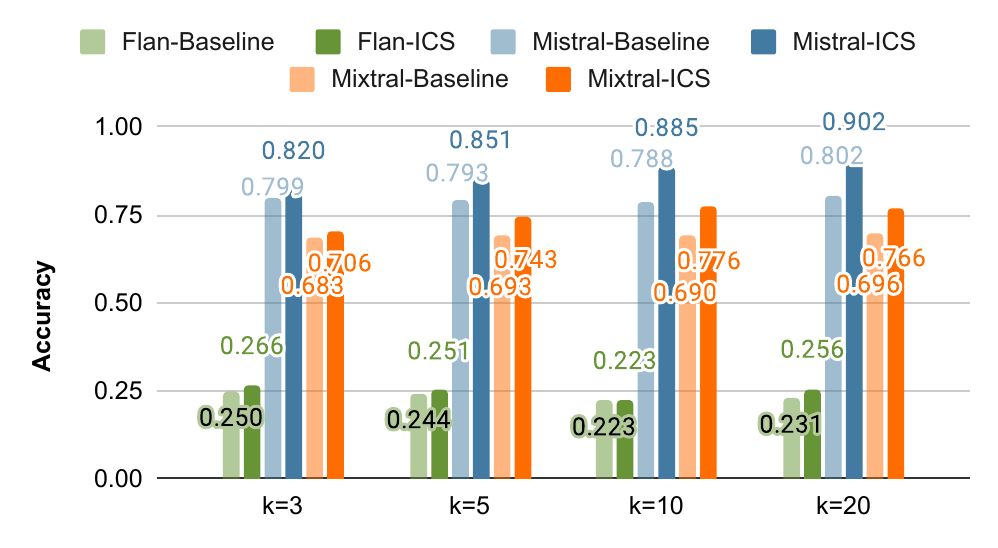}
        \caption{Contract-NLI~\cite{koreeda-manning-2021-contractnli-dataset}}
    \end{subfigure}

    \begin{subfigure}{0.495\textwidth}
        \includegraphics[width=\columnwidth]{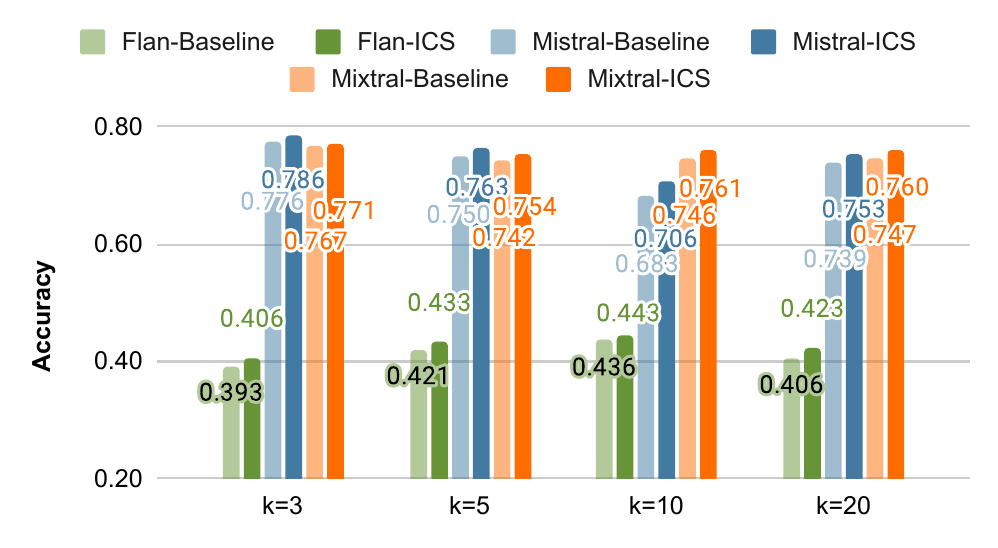}
        \caption{ CQA~\cite{talmor2018commonsenseqa} }
    \end{subfigure}

    
    



    \caption{Benchmark experiment of FLAN-T5-XL, Mistral-7B, and Mixtral-8x7B on five datasets with $100$ sampled demonstration candidates ($n$=100) for random ICS strategy compared with the baseline ICL approach. }
    \label{fig:results}
    
\end{figure*}

\subsection{Benchmark Evaluation: ICS vs. ICL}
\label{sec:evaluation-benchmark}

\subsubsection{Setup}
\label{sec:evaluation-benchmark-setup}

We conduct benchmark experiments to demonstrate the effectiveness of our ICS pipeline with a random sampling strategy for both sampling demonstration candidates and augmenting ICL prompt inputs.
The baseline setting is a traditional ICL approach with the same amount of demonstrations in each prompt input.
Specifically, we employ three open-source LLMs (FLAN-T5-XL~\cite{chung2022scaling}, Mistral-7B~\cite{jiang2023mistral}), and Mixtral-8x7B~\cite{jiang2024mixtral}, which is a Mixture-of-Experts~\cite{jacobs1991adaptive, shazeer2017outrageously} LLM. 

We experiment on three generic NLI tasks of increasing difficulties: e-SNLI~\cite{camburu2018snli}, Multi-NLI~\cite{williams2017broad}, ANLI~\cite{nie2019adversarial}, a domain-specific Contract-NLI~\cite{koreeda-manning-2021-contractnli-dataset} dataset, and the CommonsenseQA~\cite{talmor2018commonsenseqa} dataset (dataset statistics in Appendix~\ref{app:dataset-stats}). 
We originally considered Llama2~\cite{touvron2023llama2} but eventually excluded it because our preliminary experiment, discussed in Appendix~\ref{app:llama2}, shows that Llama2 tends to output ``neutral'' regardless of the inputs on ANLI. 
We also conduct a small-scale ablation study with OpenAI's close-domain GPT-3.5 in Appendix~\ref{app:gpt}.

We intended to manipulate and investigate two controlled variables of ICS: 
\textbf{the size of sampled demonstration candidates $n$}, where $n\in \{50, 100, 250, 500\}$, and \textbf{the number of augmented prompt inputs $k$ for each data to be predicted}, where $k\in \{3,5,10,20\}$.
We fix the number of demonstrations in each prompt input as three across all methodologies and experiments.
The baseline is the vanilla ICL approach with randomly chosen three examples, denoted as $baseline$ in Figure~\ref{fig:results} and $ICL$ in diagrams from Appendix~\ref{app:diagrams}.
We consider $500$ annotations a reasonable budget cap for various real-world, low-resource scenarios.  
Each setting is repeated and averaged over $10$ trials to counter the sampling randomness.
All the detailed experiment settings, including the task instruction narrative, are reported in Appendix~\ref{app:setup}.

\subsubsection{Results}
\label{sec:evaluation-benchmark-results}

The complete evaluation results for every setting are reported in Appendix~\ref{app:diagrams}.
We notice that the accuracy improvement becomes insignificant once $n$ goes beyond 100. 
This observation implies that a sample size over 100 can be considered diverse and representative enough for the tasks we experimented with, and selecting more data would have only a marginal effect on representativeness.
In Figure~\ref{fig:results}, we present the prediction accuracy of baseline ICL and our ICS strategy for every model and dataset when $n=100$.
We report the prediction accuracy as colored bars, where the \textcolor{cGREEN}{green} bars denote FLAN-T5-XL, the \textcolor{cBLUE}{blue} bars denote Mistral-7B, and the \textcolor{orange}{orange} bars denote Mixtral-8x7B.

By comparing the accuracy differences in every diagram between the baseline ICL approach and our ICS strategy for each model, we can observe that ICS can \textbf{consistently improve both LLMs' prediction performance} in every $(n, k)$ combination.
It justifies the validity of our proposed ICS paradigm.  
It is not difficult to observe that the accuracy improvement provided by the ICS strategy for FLAN-T5-XL is much less than that for Mistral-7B and Mixtral-8x7B, where the latter two models illustrate more than 5\% average improvement across all datasets with our ICS strategy.
Additionally, we observe that FLAN-T5-XL results in extremely poor performance on Contract-NLI, implying that the model lacks domain knowledge to solve this task. 
Our discussion about the potential reasons for the disparate performance between different models is detailed in Section~\ref{sec:discussion}.

\subsection{ICS Strategy Evaluation}
\label{sec:evaluation-strategy}

\subsubsection{Setup}
\label{sec:evaluation-strategy-setup}

Given the observations from the previous benchmark experiment, the best-performing ICS setting in terms of the candidate sampling size and the size of augmented prompt inputs is when $n$=100 and $k$=10.
In this ICS strategy evaluation experiment, we utilize this set of parameters and further investigate the effectiveness of different ICS strategies we introduced in Section~\ref{sec:ics-sampling} over the random ICS and baseline ICL strategies. 
We implement different ICS strategy combinations to conduct an in-depth analysis of the sampling strategies at each ICS step: sampling demonstration candidates and augmenting the prompt inputs.
We determine Mistral-7B as the backbone because it performs higher effectiveness toward ICL and more robust performance on the domain-specific dataset from the benchmark experiment, compared with FLAN-T5-XL.
Compared with Mixtral-8x7B, inferencing with Mistral-7B is faster and more cost-efficient.

\begin{table*}[!tp]
    \centering
    \resizebox{0.97\linewidth}{!}{
    \begin{booktabs}{
    colspec={llcccc},
    row{1}={font=\bfseries\small},
    width=\linewidth,
    cells={m},
}
\toprule[1pt]
{Sampling\\Strategy} & {Prompting\\Strategy} & { e-SNLI\\ \cite{camburu2018snli} } & { Multi-NLI\\ \cite{williams2017broad} } & {ANLI\\ \cite{nie2019adversarial} } & {Contract-NLI\\~\cite{koreeda-manning-2021-contractnli-dataset} } \\
\midrule
Diversity & Diversity & 73.28 $\color{cGREEN}(\uparrow 8.54)$ & 62.10 $\color{cGREEN}(\uparrow 5.20)$ & \textbf{42.78} $\color{cGREEN}(\uparrow 2.36)$ & 87.66 $\color{cGREEN}(\uparrow 8.83)$ \\
Diversity & \textit{Random} & 73.68 $\color{cGREEN}(\uparrow 8.94)$ & 62.27 $\color{cGREEN}(\uparrow 5.37)$ & 42.77 $\color{cGREEN}(\uparrow 2.35)$ & 89.42 $\color{cGREEN}(\uparrow 10.59)$ \\
\textit{Random} & Diversity & 73.47 $\color{cGREEN}(\uparrow 8.73)$ & 61.21 $\color{cGREEN}(\uparrow 4.31)$ & 42.33 $\color{cGREEN}(\uparrow 1.91)$ & 87.53 $\color{cGREEN}(\uparrow 8.70)$ \\
\midrule
Similarity & Similarity & 73.63 $\color{cGREEN}(\uparrow 8.89)$ & 61.79 $\color{cGREEN}(\uparrow 4.89)$ & 42.47 $\color{cGREEN}(\uparrow 2.05)$ & 90.44 $\color{cGREEN}(\uparrow 11.61)$ \\
Similarity & \textit{Random} & \textbf{74.11} $\color{cGREEN}(\uparrow 9.37)$ & 62.09 $\color{cGREEN}(\uparrow 5.19)$ & 42.60 $\color{cGREEN}(\uparrow 2.18)$ & \textbf{90.48} $\color{cGREEN}(\uparrow 11.65)$ \\
\textit{Random} & Similarity & 73.74 $\color{cGREEN}(\uparrow 9.00)$ & 62.17 $\color{cGREEN}(\uparrow 5.27)$ & 42.63 $\color{cGREEN}(\uparrow 2.21)$ & 88.88 $\color{cGREEN}(\uparrow 10.05)$  \\
\midrule
Hybrid & Hybrid & 73.86 $\color{cGREEN}(\uparrow 9.12)$ & \textbf{62.52} $\color{cGREEN}(\uparrow 5.62)$ & 42.59 $\color{cGREEN}(\uparrow 2.17)$ & 88.85 $\color{cGREEN}(\uparrow 10.02)$ \\
Hybrid & \textit{Random} & 73.96 $\color{cGREEN}(\uparrow 9.22)$ & 62.41 $\color{cGREEN}(\uparrow 5.51)$ & 42.56 $\color{cGREEN}(\uparrow 2.14)$ & 89.73 $\color{cGREEN}(\uparrow 11.90)$ \\
\textit{Random} & Hybrid & 73.95 $\color{cGREEN}(\uparrow 9.21)$ & 62.39 $\color{cGREEN}(\uparrow 5.49)$ & 42.45 $\color{cGREEN}(\uparrow 2.03)$ & 89.06 $\color{cGREEN}(\uparrow 10.23)$ \\
\midrule
\textit{Random} & \textit{Random} & 72.57 $\color{cGREEN}(\uparrow 7.83)$ & 61.17 $\color{cGREEN}(\uparrow 4.27)$ & 42.22 $\color{cGREEN}(\uparrow 1.80)$ & 86.69 $\color{cGREEN}(\uparrow 7.86)$ \\
\midrule[1pt]
\SetCell[c=2]{c}{ICL (Baseline)} && 64.742 & 56.905 & 40.420 & 78.83 \\

\bottomrule[1pt]
\end{booktabs}}
\caption{
Comparison of different ICS strategies versus the ICL baseline on four datasets with Mistral-7B~\cite{jiang2023mistral}. We implement different strategy combinations and average each score over $40$ trials. 
The change in prediction accuracy compared with the traditional ICL approach is reported in the parenthesis.} 
\label{tab:ics-strategy-evaluation}
\end{table*}

Because of the massive size of e-SNLI and Multi-NLI ($540$k and $390$k in train splits, correspondingly), we borrow the concept from Active Learning simulations~\cite{yao2023labels} to efficiently evaluate the strategies with a reasonable amount of data and acquire the averaged score over multiple trials.
Specifically, for each trial, we randomly sample $3,000$ and $1,000$ data from the train and test split correspondingly as the actual train and test data for the current trial.
We then conduct each setting $40$ trials to minimize the randomness provided by subsampling training and testing data and report the averaged prediction accuracy in Table~\ref{tab:ics-strategy-evaluation}.

\subsubsection{Results}
\label{sec:evaluation-strategy-results}

In addition to the prediction accuracy of different ICS strategy combinations, we also report the change in prediction accuracy compared with the baseline ICL approach in the parenthesis, where \textcolor{cGREEN}{green} denotes improvement. 
We can easily observe that all three ICS sampling strategies (diversity, similarity, and hybrid) can \textbf{consistently and significantly improve the prediction accuracy of Mistral-7B} compared with the baseline setting, with more than $9\%$ improvement on e-SNLI and two-digits elevation on Contract-NLI.
It is worth noticing that all the ICS settings with non-random strategies in at least one ICS step can outperform the benchmark ICS setting that utilizes the random strategy for both sampling and prompt augmentation.
From the results, we can also observe that no single best strategy exists, even for the same NLI task.
This observation is aligned with our motivation and the aforementioned existing works that different ICL demonstrations provide distinct knowledge about the task, and there's no single best ICL strategy yet. 
Specifically, the diversity strategy stands out on ANLI, whereas the hybrid strategy outperforms the other strategies on Multi-NLI, and the similarity strategy surpasses the others on e-SNLI as well as Contract-NLI.

Additionally, we observe that non-random strategies do not lead to consistent performance improvement for augmenting ICL prompt inputs by comparing them with the random strategy.
For example, leveraging the random strategy for augmenting prompt inputs outperforms the similarity strategy on all four datasets, implying that \textbf{high similarity among the demonstrations within each prompt input is not preferred}. 
On the other hand, we can observe a significant performance improvement in leveraging non-random strategies demonstration candidate sampling compared to the random strategy.
This observation leads to the conclusion that all three strategies demonstrate more contributions during demonstration candidate sampling compared with augmenting ICL prompt inputs.
We also hypothesize that more carefully curated strategies are needed to sample ICL combinations effectively, leaving a broader avenue for future research.

Furthermore, we notice \textbf{the improvement provided by ICS sampling strategies is inversely proportional to the difficulty of the tasks}.
If we consider the model's baseline ICL performance from Section~\ref{sec:evaluation-benchmark} as a faithful indicator of dataset difficulty, we can conclude that the dataset ordering in ascending order of task difficulty will be e-SNLI, Multi-NLI, and ANLI, where the performance improvement provided by ICS strategies is the smallest on ANLI and the largest on e-SNLI.

Our evaluation of different ICS strategies illustrates promising results that fundamental similarity-based algorithms can effectively increase ICS enhancement, leading to broader future research avenues in exploiting the benefits of more carefully curated ICS strategies with LLMs.

\section{Discussion}
\label{sec:discussion}

\paragraph{Limited Performance with FLAN-T5} 
FLAN-T5 models have been fine-tuned on various downstream tasks, including NLI. This fine-tuning could indeed influence the models' performance in in-context learning scenarios, potentially skewing the effectiveness of ICS.
Additionally, we observe FLAN-T5-XL results in poor performance on Contract-NLI from Figure~\ref{fig:results}, despite it can perform adequately well on the other three generic-domain NLI datasets. 
We conduct an ablation study with FLAN-T5-XL for ICL to investigate the potential reasons and report in Appendix~\ref{app:flant5}.
Given the ablation study results, we hypothesize several possible reasons: 1) FLAN-T5-XL falls short of properly interpreting long text sequences; 2) FLAN-T5-XL was not fine-tuned to elevate the ability to interpret ICL demonstrations, and 3) FLAN-T5-XL lacks the necessary domain knowledge to solve the Contract-NLI task.

\paragraph{ICS-Related Work} A very recent work attempts multiple ICL methodologies to investigate whether LLMs can beat domain-specific fine-tuned models in the medical domain~\cite{nori2023can}.
The \textit{Choice Shuffling Ensemble} technique in their proposed ensemble methodology shares a similar concept with our proposed ICS paradigm, but the authors only focus on shuffling the answer choices for selecting robust predictions. 
Nevertheless, we believe that ICS depicts vast prospects and potential to exploit the capabilities of LLMs.

\section{Conclusion}
\label{sec:conclusion}

This work presents In-context Sampling (ICS), a novel In-Context Learning paradigm for probing confident predictions by sampling demonstration candidates and augmenting different ICL prompt inputs.
Our experiments show that even ICS with the random strategy can lead to consistent accuracy improvement compared with the traditional ICL approach, and further illustrate the additional helpfulness provided by three fundamental but effective data similarity-based sampling strategies with ICS.
Our work lays the foundation for implementing ICL-based applications to support non-expert users in the real world, as they do not know how to write a single perfect prompt to get their work done but often write multiple prompt inputs~\cite{10.1145/3544548.3581388}. 
Our method aligns well with such user scenarios.

\section{Limitations}
\label{sec:limitation}


The primary focus of this paper is to propose and demonstrate the effectiveness of our ICS pipeline compared with the traditional ICL approach. 
Thus, we do not compare with other prompting strategies that do not focus on in-context demonstrations, such as Chain-of-Thoughts. 
Our experiments showed that ICS can improve the model's performance (in prediction accuracy) even with a random strategy. 
We further illustrate the potential of three proposed similarity-based ICS strategies, which, despite fundamental, can further exploit LLM's capability and boost the prediction performance.

However, despite extensive experiments with different $n$ and $k$ combinations, several potential variables require further analysis. 
For instance, we considered five datasets of different difficulties and each ICL combination is arbitrary, where four of the datasets are NLI tasks and the other one is a commonsense QA task. 
The generalizability of the ICS paradigm to other types of tasks goes beyond the scope of this paper, and we are working on this interesting and substantial research question as a follow-up work, especially in real-world scenarios. 

We only implement and evaluate the same three strategies for both steps of sampling demonstration candidates and augmenting prompt inputs in ICS because the data similarity-based strategies are model agnostic and generally require fewer computing resources than model-based strategies. 
We are also aware that the optimal strategy for demonstration candidate sampling may not be optimal for prompt input augmentations, and we leave the analysis of strategy optimization for future work.

In addition, we do not perform an in-depth analysis of optimizing time consumption and reducing computing resources in this work, though we are aware that ICS may require more time than the traditional ICL approach. 
Lastly, our experiment comprises four open-source LLMs as the original plan but excludes Llama2 due to its over inclination to predict the ``neutral'' category (Appendix~\ref{app:llama2}). 
We identify that there are still a variety of other instructional-finetuned LLMs we do not include in this work, such as InstructGPT~\cite{ouyang2022training}. 
We do not focus on close-sourced and commercial-oriented LLMs such as GPT-4~\cite{openai2023gpt4} in this work. However, we report a small-scale ablation study with GPT-3.5 in Appendix~\ref{app:gpt} that further illustrates the generalizability of our proposed ICS strategy. 



\bibliography{anthology,custom,yuxuan}

\clearpage
\appendix

\section{Experiment Setup}
\label{app:setup}


We incorporate four natural language inference datasets (e-SNLI, Multi-NLI, ANLI, and Contract-NLI) in our evaluation. 
Thus, we leverage the same instruction narrative across all the experiments for these datasets: \textbf{Determine whether a hypothesis is \textit{entailment, neutral, contradiction} giving a premise.}
For Contract-NLI, the original dataset only consists of annotations for the ``entailment'' and ``contradiction'' categories. Thus, we only evaluate the performance of those data. 
For CommonsenseQA, we design the prompt to be: \textbf{Answer this commonsense question from the given choices.}

All the experiments are computed on one of two resources: 1) an NVIDIA A100 40G graphic card or 2) an NVIDIA 3090 24G graphic card.
To fit the models in both graphic cards, we load both Llama2 and Mistral-7B in fp16 precision, load Mixtral-8x7B in 4-bit precision, and limit to generate a maximum of 10 tokens.

\section{Dataset Statistics}
\label{app:dataset-stats}

\begin{table}[!t]
  \center
  \small
  \begin{booktabs}{
    width=\columnwidth,
    colspec={X[l]ccc},
    hspan=minimal,
    cells={m}
  }
    \toprule
    Dataset & Train & Validation & Test \\
    \midrule
    {e-SNLI \\ {\scriptsize \citet{camburu2018snli}}} & $549,367$ & $9,842$ & $9,824$ \\
    {Multi-NLI \\ {\scriptsize \citet{williams2017broad}}} & $392,702$ & $9,815$ & $9,832$ \\
    {ANLI \\ {\scriptsize \citet{nie2019adversarial}}} & $16,946$ & $1,000$ & $1,000$ \\
    {Contract-NLI \\ {\scriptsize \citet{koreeda-manning-2021-contractnli-dataset}}} & $3,999$ & $555$ & $1,113$ \\
    {CommonsenseQA \\ {\scriptsize \citet{talmor2018commonsenseqa}}} & $9,741$ & $1,221$ & $1,140$ \\

    \bottomrule
  \end{booktabs}
  \caption{Datasets involved in our experiment. Contract-NLI only comprises annotations of ``entailment'' and ``contradiction'' categories.}
  \label{tab:dataset-stats}
\end{table}

\section{Complete Evaluation Results}
\label{app:diagrams}

\begin{figure*}[!ht]
    \centering
    \subfloat[n=50]{
    \includegraphics[width=0.98\columnwidth]{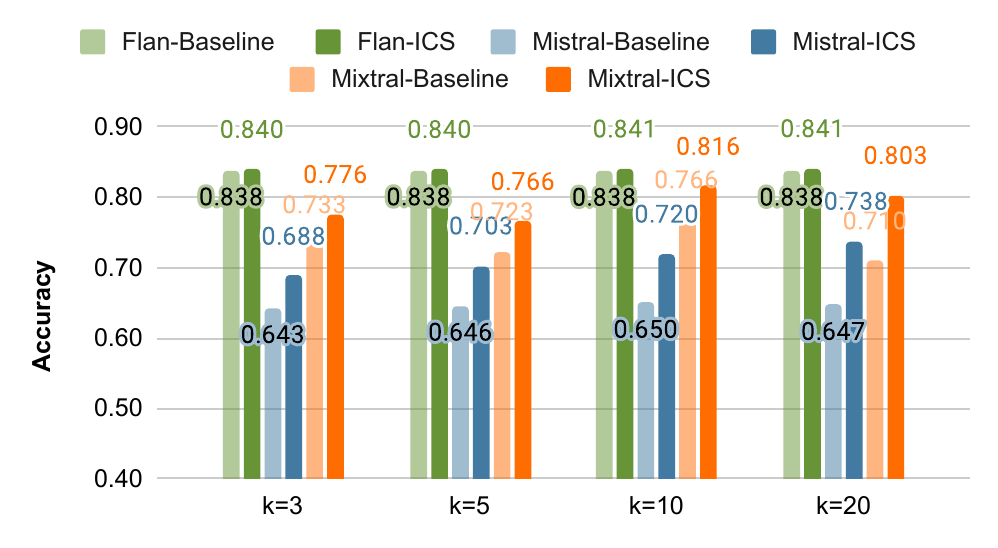}
    }
    \hfill
    \subfloat[n=100]{
        \includegraphics[width=0.98\columnwidth]{figures/esnli_n=100.pdf}
    }

    \medskip

    \subfloat[n=250]{
    \includegraphics[width=0.98\columnwidth]{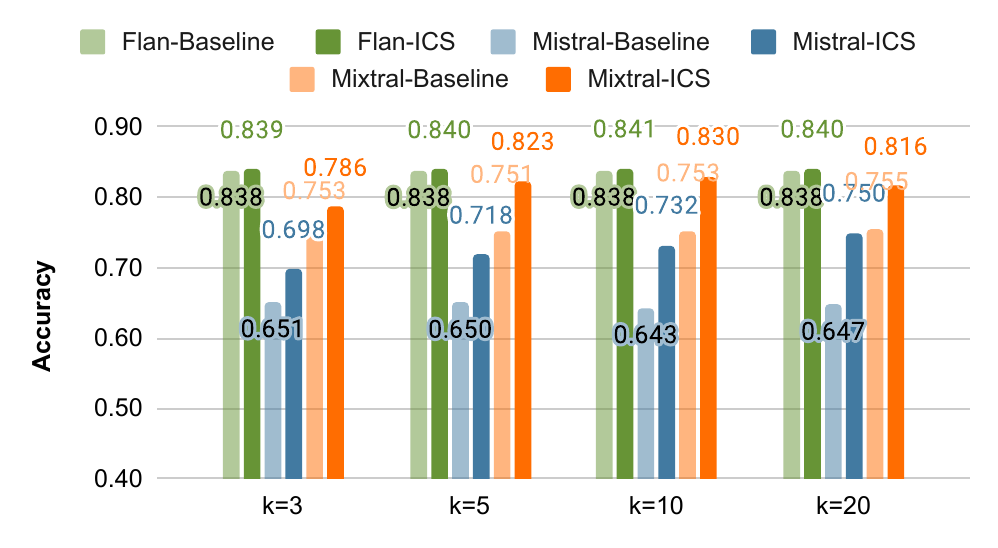}
    }
    \hfill
    \subfloat[n=500]{
        \includegraphics[width=0.98\columnwidth]{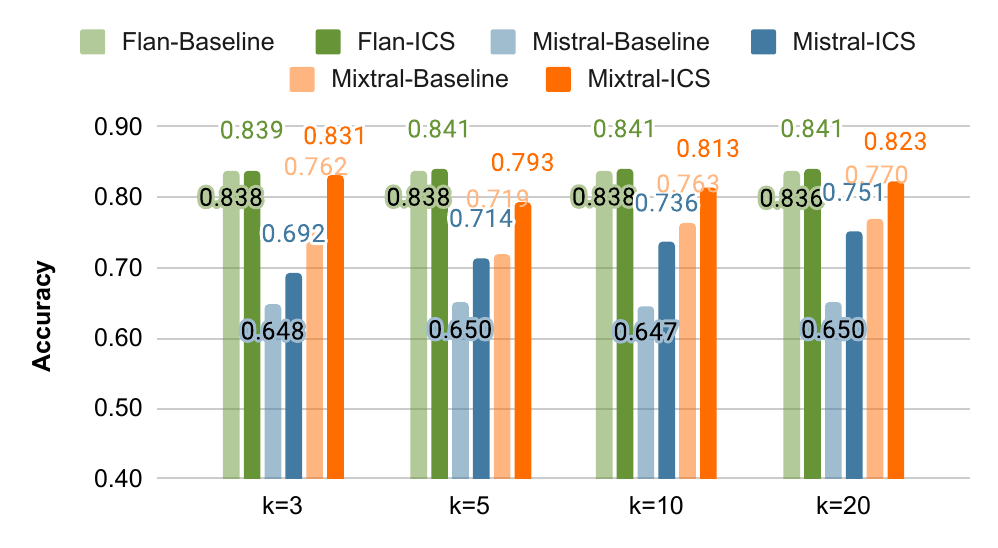}
    }

    \caption{ Evaluation results with FlanT5-XL, Mistral, and Mixtral on e-SNLI~\cite{camburu2018snli} dataset.}
    
    \label{fig:result_esnli}
\end{figure*}

\begin{figure*}[!ht]
    \centering
    \subfloat[n=50]{
    \includegraphics[width=0.98\columnwidth]{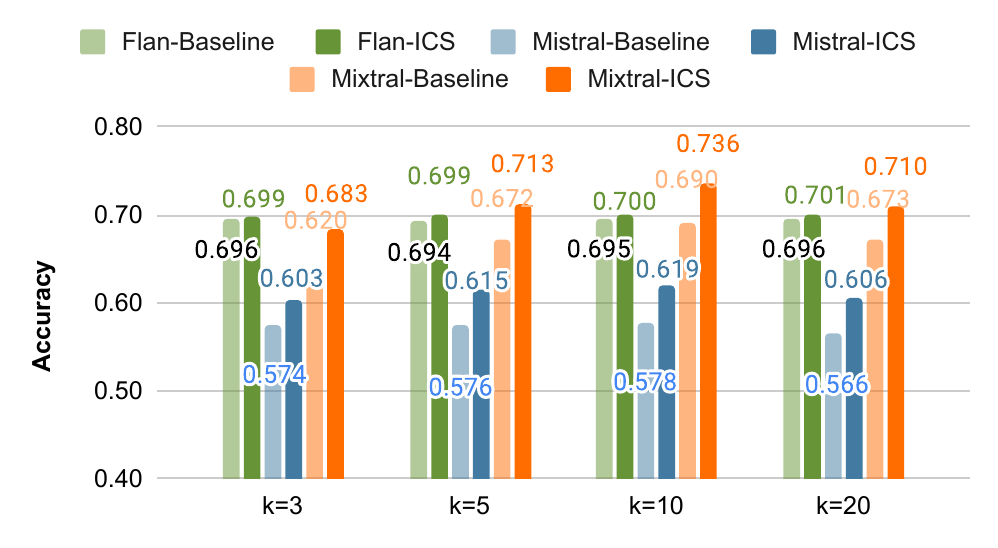}
    }
    \hfill
    \subfloat[n=100]{
        \includegraphics[width=0.98\columnwidth]{figures/multinli_n=100.pdf}
    }

    \medskip

    \subfloat[n=250]{
    \includegraphics[width=0.98\columnwidth]{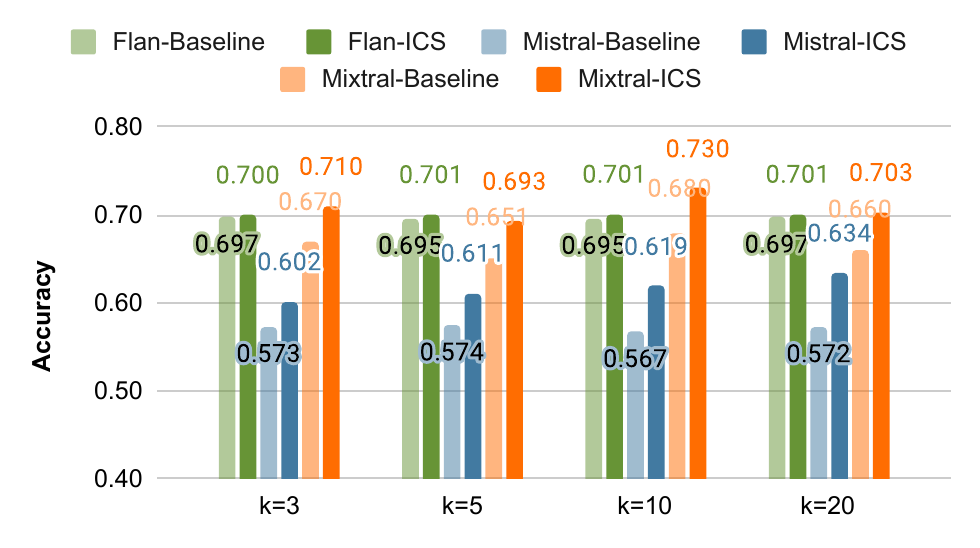}
    }
    \hfill
    \subfloat[n=500]{
        \includegraphics[width=0.98\columnwidth]{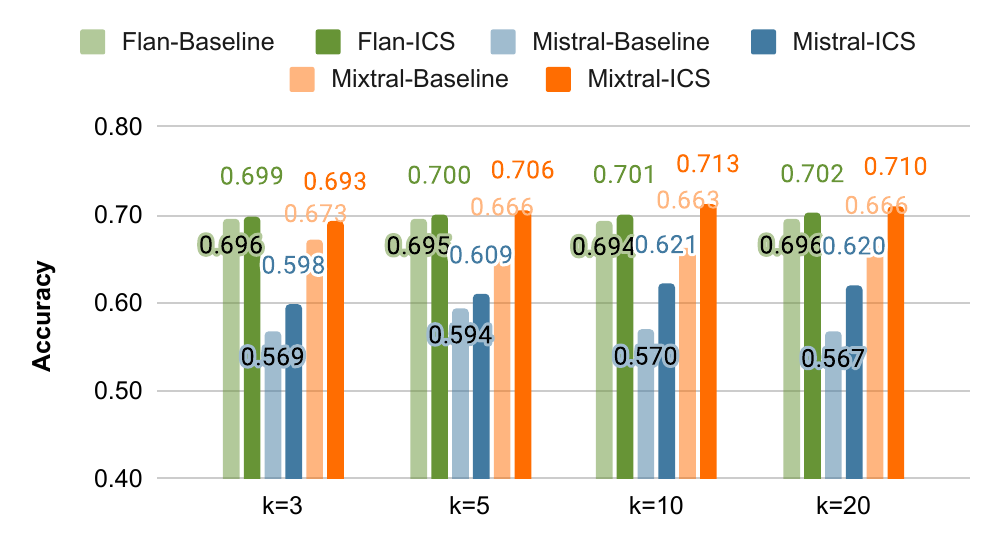}
    }

    \caption{ Evaluation results with FlanT5-XL, Mistral, and Mixtral on Multi-NLI~\cite{williams2017broad} dataset.}
    
    \label{fig:result_multinli}
\end{figure*}

\begin{figure*}[!ht]
    \centering
    \subfloat[n=50]{
    \includegraphics[width=0.98\columnwidth]{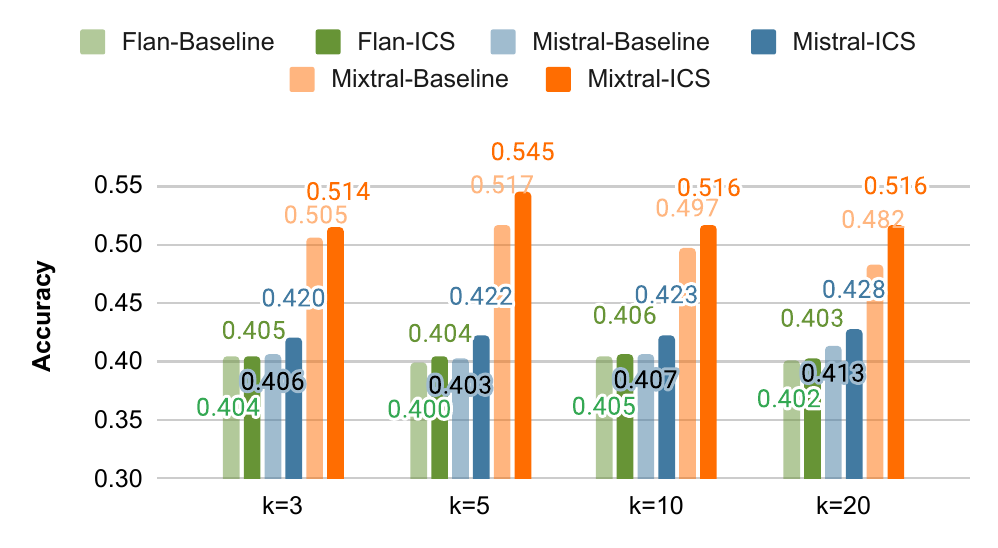}
    }
    \hfill
    \subfloat[n=100]{
        \includegraphics[width=0.98\columnwidth]{figures/ANLI_n=100.pdf}
    }

    \medskip

    \subfloat[n=250]{
    \includegraphics[width=0.98\columnwidth]{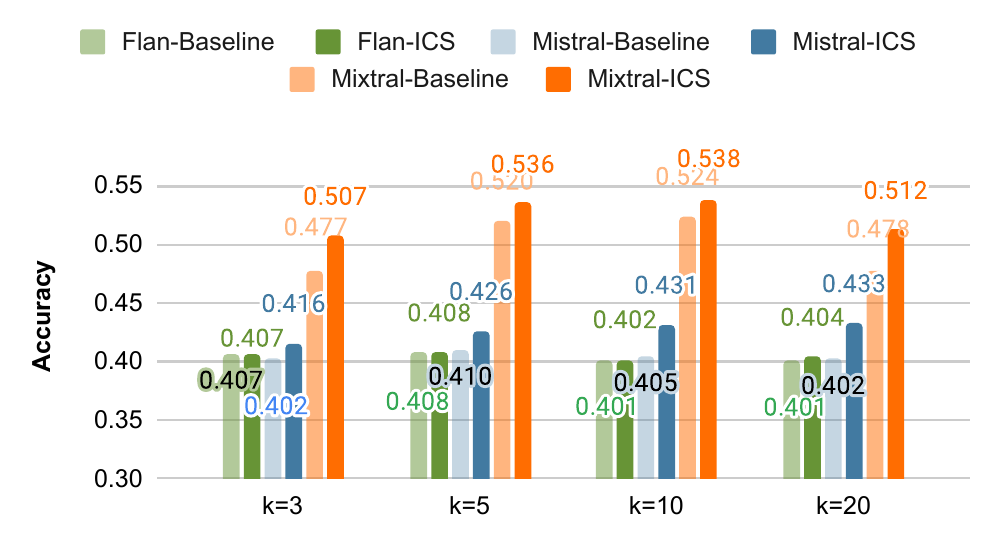}
    }
    \hfill
    \subfloat[n=500]{
        \includegraphics[width=0.98\columnwidth]{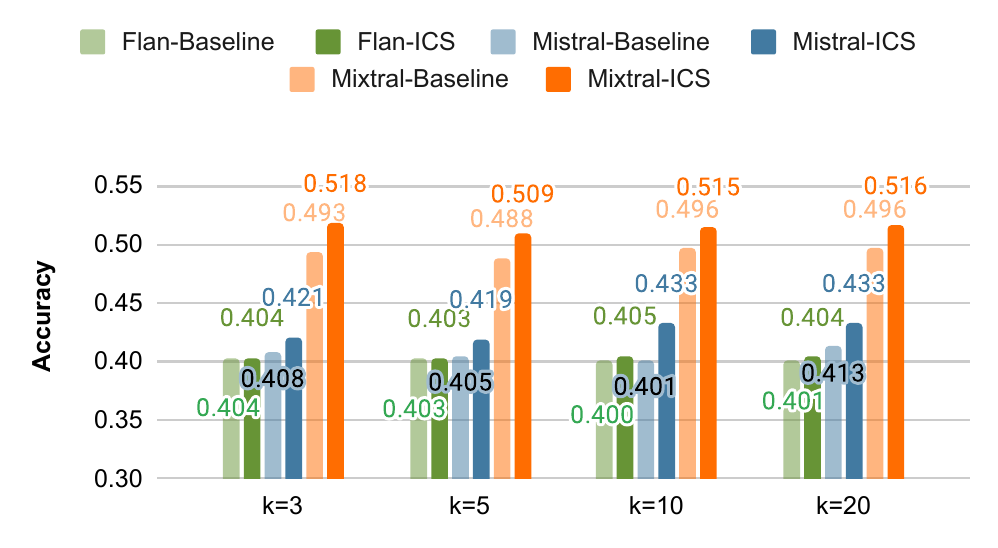}
    }

    \caption{ Evaluation results with FlanT5-XL, Mistral, and Mixtral on ANLI~\cite{nie2019adversarial} dataset.}
    
    \label{fig:result_anli}
\end{figure*}

\begin{figure*}[!ht]
    \centering
    \subfloat[n=50]{
    \includegraphics[width=0.98\columnwidth]{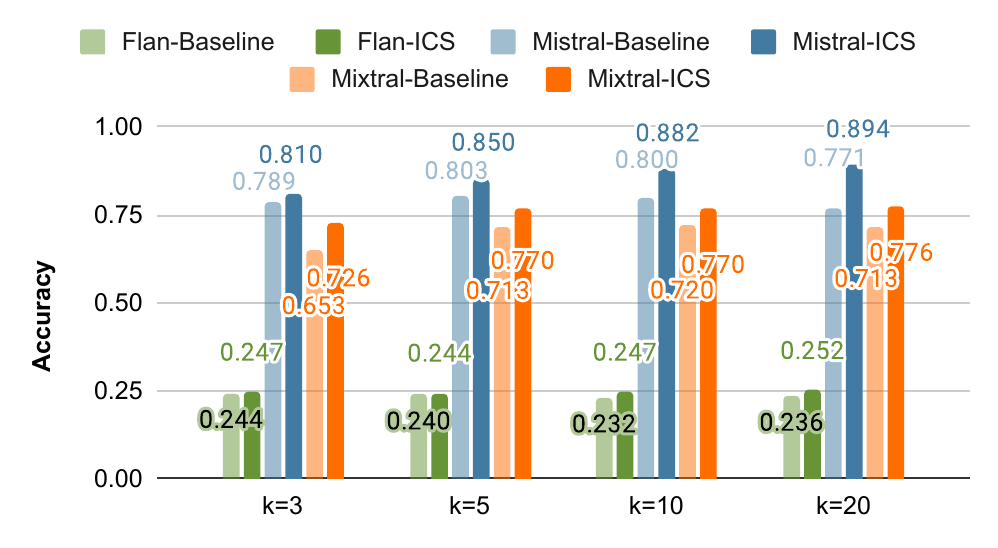}
    }
    \hfill
    \subfloat[n=100]{
        \includegraphics[width=0.98\columnwidth]{figures/contractnli_n=100.pdf}
    }

    \medskip

    \subfloat[n=250]{
    \includegraphics[width=0.98\columnwidth]{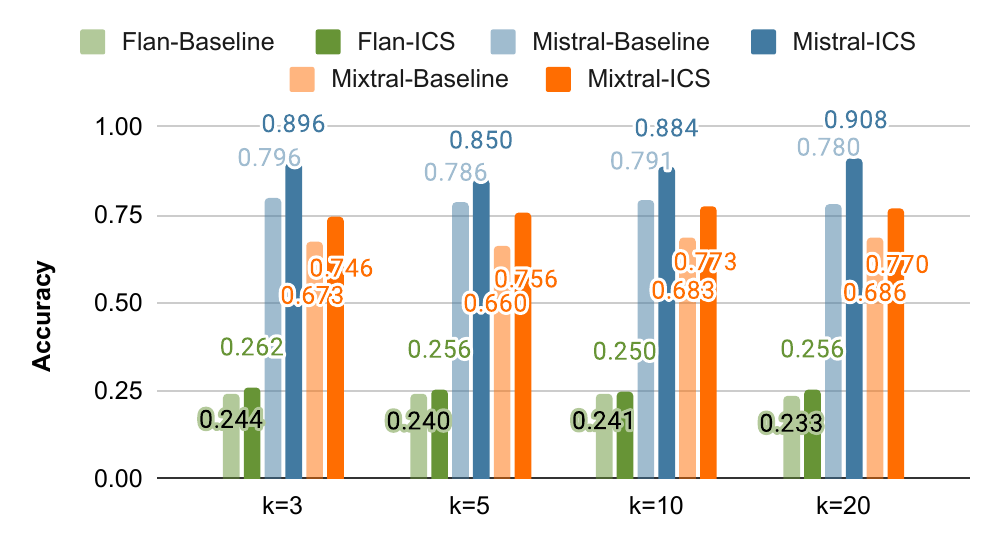}
    }
    \hfill
    \subfloat[n=500]{
        \includegraphics[width=0.98\columnwidth]{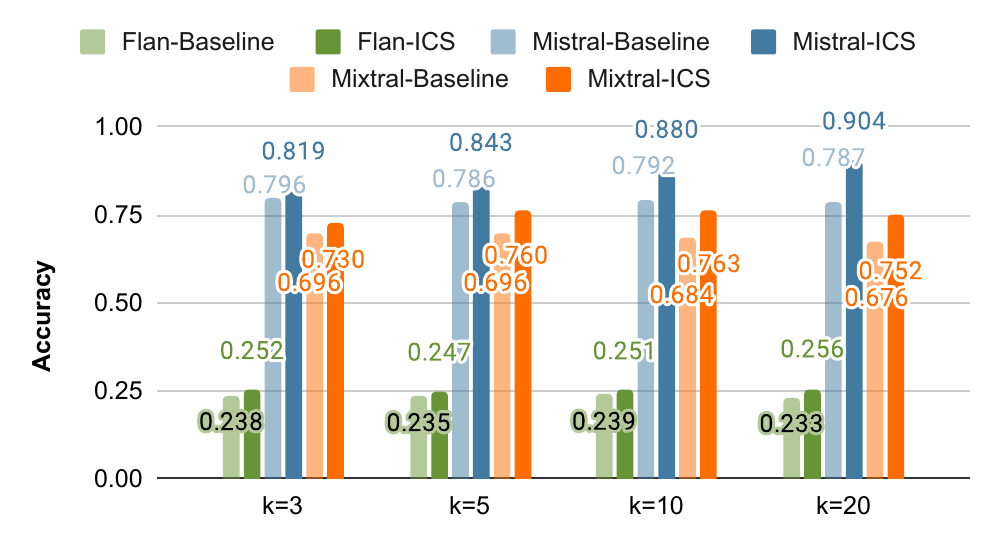}
    }

    \caption{ Evaluation results with FlanT5-XL, Mistral, and Mixtral on Contract-NLI~\cite{koreeda-manning-2021-contractnli-dataset} dataset.}
    
    \label{fig:result_contractnli}
\end{figure*}

\begin{figure*}[!ht]
    \centering
    \subfloat[n=50]{
    \includegraphics[width=0.98\columnwidth]{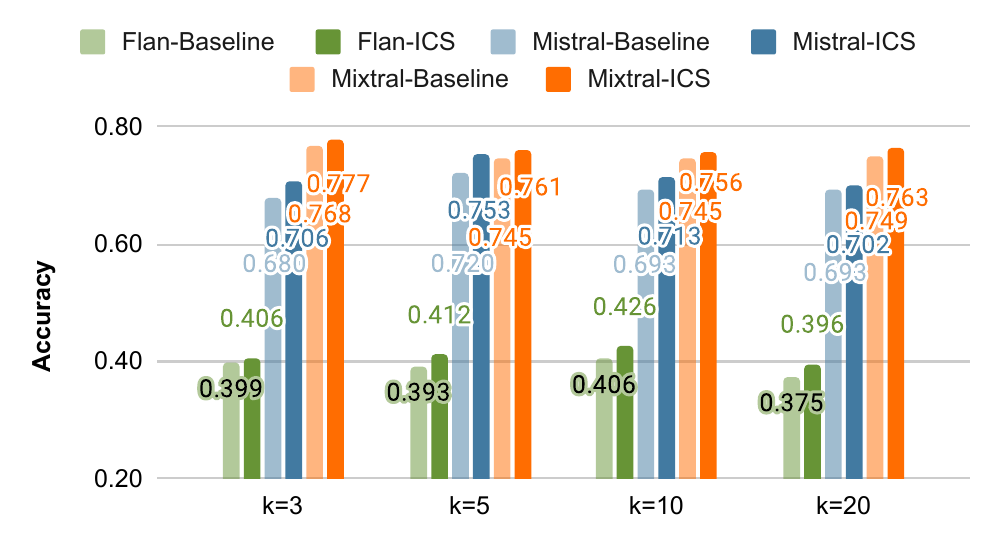}
    }
    \hfill
    \subfloat[n=100]{
        \includegraphics[width=0.98\columnwidth]{figures/CQA_n=100.pdf}
    }

    \medskip

    \subfloat[n=250]{
    \includegraphics[width=0.98\columnwidth]{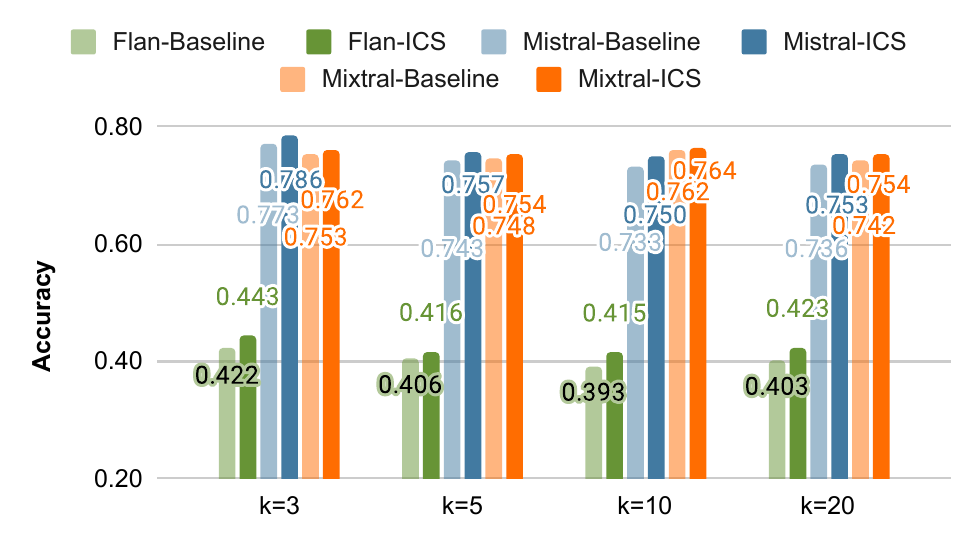}
    }
    \hfill
    \subfloat[n=500]{
        \includegraphics[width=0.98\columnwidth]{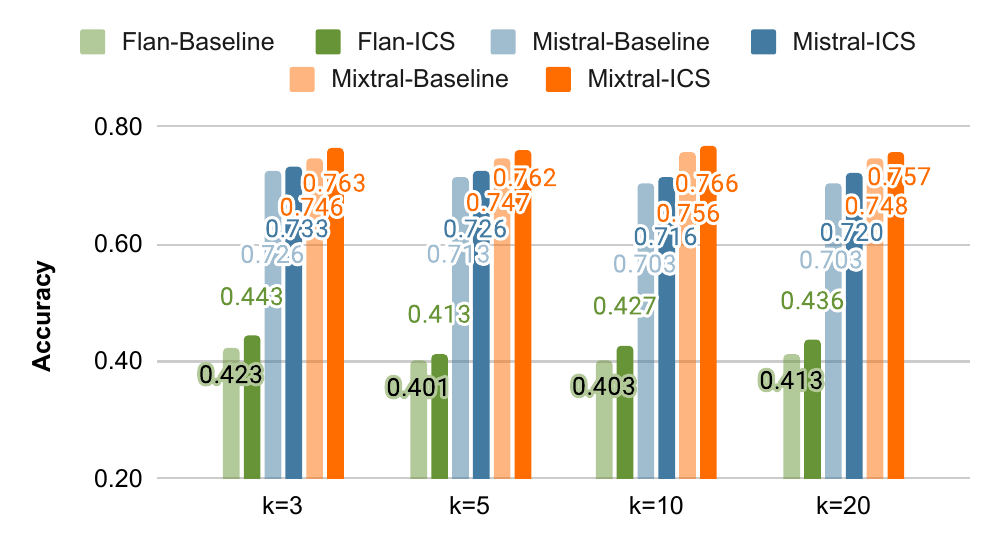}
    }

    \caption{ Evaluation results with FlanT5-XL, Mistral, and Mixtral on CQA~\cite{talmor2018commonsenseqa} dataset.}
    
    \label{fig:result_cqa}
\end{figure*}

Here, we report the complete results of our evaluation (Section~\ref{sec:evaluation}) in Figure~\ref{fig:result_esnli},~\ref{fig:result_multinli},~\ref{fig:result_anli},~\ref{fig:result_contractnli},~\ref{fig:result_cqa} on e-SNLI, Multi-NLI, ANLI, Contract-NLI, and CQA, correspondingly.
We acquire an average prediction accuracy score over $10$ trials of each setting. $n$ denotes the amount of demonstration candidate data we sampled, and $k$ denotes the number of ICL combinations for each test data.


We can observe that the ICS strategy can consistently improve LLMs' performance compared with the traditional ICL baseline; in addition, FLAN-T5-XL is much less sensitive than Mistral and Mixtral toward the improvement provided by the ICS strategy.
From the diagrams, $k=10$ and $n=100$ are the best-performing parameters that maximize the performance improvement and minimize the standard deviations.

\section{Analysis on Llama2}
\label{app:llama2}

\begin{table}[!t]

\centering
\small
\resizebox{\columnwidth}{!}{
\begin{tabular}{@{}llcccc@{}}
\toprule

\multirow{2}{*}{\textbf{Llama2}}                          & \multirow{2}{*}{\textbf{Inst. 1}} & \multirow{2}{*}{\textbf{Inst. 2}} & \multirow{2}{*}{\textbf{Inst. 3}} & \multirow{2}{*}{\textbf{Ground-truth}}    \\
\multicolumn{2}{l}{}        &           &           &           &     \\ \midrule

\textbf{entailment}         &  75       &  202      &  151      &   334       \\
\textbf{neutral}            &   808     &  668      &  785      &  333       \\
\textbf{contradiction}      &  117      &  130      &  64       &    333       \\

\bottomrule
\end{tabular}
}
\caption{Analysis of Llama2 performance on ANLI.}
\label{tab:llama2_anli}
\end{table}

We conduct an initial inference experiment with Llama2~\cite{touvron2023llama2} on ANLI utilizing three different natural language instructions: 
\begin{enumerate}[label=\roman*,nosep]
    \item Determine whether a hypothesis is \textit{entailment, neutral, contradiction} giving a premise.
    \item Classifying a pair of premise and hypothesis sentences into three classes: \textit{entailment, neutral, contradiction}
    \item Predict the relationship between the premise and hypothesis by \textit{entailment, neutral, contradiction}
\end{enumerate}
The results are reported in Table~\ref{tab:llama2_anli}. 
We can easily observe that Llama2 tends to overly predict ``neutral'' over the other two categories despite changing instruction narratives, whereas the ground-truth distribution is even across categories. 
Thus, we omit Llama2 in our work. 
There could be different reasons contributing to this issue; for example, Llama2 was overfitted to the NLI task or similar tasks that share the same set of targeting categories: ``entailment'', ``neutral'', and ``contradiction''.

\begin{table}[!t]
    \centering
    \resizebox{\columnwidth}{!}{
    \begin{booktabs}{
    colspec={lcccc},
    row{1}={font=\bfseries\small},
    width=\columnwidth,
    cells={m},
}
\toprule[1pt]
{Setting} & {e-SNLI} & {Multi-NLI} & {ANLI} & {CommonsenseQA}  \\
ICL & 0.57 & 0.55 & 0.55 & 0.78 \\
ICS & \textbf{0.59} & \textbf{0.6} & \textbf{0.58} & \textbf{0.81} \\
\bottomrule[1pt]
\end{booktabs}}
\caption{Ablation study with GPT-3.5 on four datasets. }
\label{tab:app-gpt}
\end{table}

\section{Ablation Study with GPT-3.5}
\label{app:gpt}

We extend the scope of our work by conducting ablation experiments with OpenAI's close-domain GPT-3.5 on four datasets. 
For each dataset, we randomly sample 200 examples from the test split and report the averaged accuracy on three trials, due to budget limit. 
Our result in Table~\ref{tab:app-gpt} shows that the proposed ICS strategy can consistently improve the performance of close-domain LLMs as well, strengthening the generalizability of our strategy.

\section{Ablation on FLAN-T5-XL with Contract-NLI}
\label{app:flant5}

\begin{table}[!tp]
    \centering
    \resizebox{\columnwidth}{!}{
    \begin{booktabs}{
    colspec={lcccc},
    row{1}={font=\bfseries\small},
    width=\columnwidth,
    cells={m},
}
\toprule[1pt]
{Setting} & {zero-shot} & {1-shot} & {2-shot} & {3-shot}  \\
ICL & 2.48 & 19.39 & 23.80 & 22.88 \\
ICS & / & 20.03 & 24.54 & 23.34 \\
\bottomrule[1pt]
\end{booktabs}}
\caption{ICL ablation experiment of FLAN-T5-XL on Contract-NLI. }
\label{tab:discussion-flant5}
\end{table}

We design and conduct an ablation study with FLAN-T5-XL for ICL to verify our hypothesis.
The experiment is conducted on the Contract-NLI dataset.
Specifically, we start with the zero-shot setting to examine whether FLAN-T5-XL can properly solve the task without demonstrations. 
Then, we experiment with both ICS and ICL approaches and gradually increase the number of demonstrations from 1 to 3. 
The demonstrations are randomly selected from the training split, and each ICL setting is repeated 3 times to acquire the average score.
From table~\ref{tab:discussion-flant5}, we can observe that FLAN-T5-XL can hardly interpret the dataset and solve it with a zero-shot setting.
Since we leverage the same prompt narrative as the one for the other NLI tasks that FLAN-T5-XL performs relatively well, we can imply that the lack of domain knowledge might be the primary reason for such low performance. 
Nevertheless, we can observe that the 1-shot setting can significantly improve the model performance, although the overall accuracy is still very low.
It is worth noticing that the improvement becomes relatively trivial once we add more demonstrations to the prompt inputs, which implies that FLAN-T5-XL falls short of interpreting longer and more complex ICL format, possibly due to its relatively short training input length limit. 
Moreover, our random ICS strategy can still outperform the ICL baseline across all settings.

\clearpage

\end{document}